\documentclass[conference]{IEEEtran}
\IEEEoverridecommandlockouts
\usepackage{amsmath,amssymb,amsfonts}
\usepackage{algorithm}
\usepackage{hyperref}
\usepackage{comment}
\usepackage{caption}
\usepackage{algpseudocode}
\usepackage{graphicx}
\graphicspath{{./images/}}
\usepackage{textcomp}
\usepackage{xcolor}
\def\BibTeX{{\rm B\kern-.05em{\sc i\kern-.025em b}\kern-.08em
    T\kern-.1667em\lower.7ex\hbox{E}\kern-.125emX}}
\usepackage
[maxbibnames=99,
firstinits=true,
backend=biber]
{biblatex}

\hypersetup{
    colorlinks=true,
    linkcolor=black,
    filecolor=black,
    citecolor=black,
    urlcolor=blue,
    }
\begin{document}

\title{Towards Computer-Vision Based Vineyard Navigation for Quadruped Robots\\
}

\author{\IEEEauthorblockN{Lee Milburn}
\IEEEauthorblockA{\textit{Dynamic Legged Systems Lab} \\
\textit{Istituto Italiano di Tecnologia}\\
Genova, Italy \\
lee.milburn@iit.it}

\and
\IEEEauthorblockN{Juan Gamba}
\IEEEauthorblockA{\textit{Dynamic Legged Systems Lab} \\
\textit{Istituto Italiano di Tecnologia}\\
Genova, Italy \\
juan.gamba@iit.it}
\and
\IEEEauthorblockN{Claudio Semini }
\IEEEauthorblockA{\textit{Dynamic Legged Systems Lab} \\
\textit{Istituto Italiano di Tecnologia}\\
Genova, Italy \\
claudio.semini@iit.it}

}

\maketitle

\begin{abstract}
There is a dramatic shortage of skilled labor for modern vineyards. The Vinum project is developing a mobile robotic solution to autonomously navigate through vineyards for winter grapevine pruning. This necessitates an autonomous navigation stack for the robot pruning a vineyard. The Vinum project is using the quadruped robot HyQReal. This paper introduces an architecture for a quadruped robot to autonomously move through a vineyard by identifying and approaching grapevines for pruning. The higher level control is a state machine switching between searching for destination positions, autonomously navigating towards those locations, and stopping for the robot to complete a task. The destination points are determined by identifying grapevine trunks using instance segmentation from a Mask Region-Based Convolutional Neural Network (Mask-RCNN). These detections are sent through a filter to avoid redundancy and remove noisy detections. The combination of these features is the basis for the proposed architecture.
\end{abstract}


\section{Introduction}

\begin{figure}[h]
\centering
\includegraphics[width=0.48\textwidth]{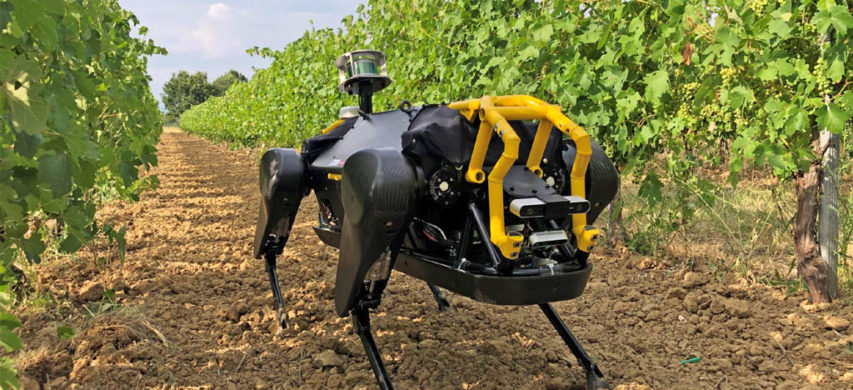}
\caption{HyQReal in Vineyard.}
\label{fig:Vinum}
\end{figure}

There is a major shortage of labor in vineyards across the world. Vineyards rely on seasonal labor, which in a lot of cases includes international workforces. Seasonal labor shortages began with the COVID-19 pandemic and have continued since\footnote[1]{https://www.winemag.com/2021/12/07/wine-industry-labor-supply/}.  Vineyards have looked towards robotic automation of seasonal work to account for the labor shortage.

The Vinum project is built on the HyQReal quadruped robot that is being developed to autonomously do the winter pruning of grapevines, see  Fig. \ref{fig:Vinum} \cite{hyqreal}. To accomplish this, the Vinum robot has to autonomously navigate vineyards, arriving at each grapevine that needs winter pruning. This extended abstract introduces a navigation architecture based on computer vision for quadruped robots. Previous vineyard navigation has described moving down each row, using a laser sensor, until there are no more grapevines in a row \cite{inproceedings}.  Other navigation stacks have been developed which also move down rows but they use laser scanners for perception \cite{rob_farmer}. Our proposed navigation stack initializes itself with a search of a vineyard row and will choose whether to start from right or left. It uses computer vision to detect the grapevines and a filter to average the detections and eliminate noise. In other papers, grapevine trunks were identified using instance segmentation \cite{loop_closure}. We implemented a similar sensor navigation control using a RGB-D for grapevine trunk image segmentation. So, detections of the grapevine trunks are made using a Mask-RCNN trained off a created dataset with 100 images. The combination of the higher level control with the grapevine detections makes the basis for the Vinum navigation stack.

The contribution of this extended abstract is a navigation for precise placement of quadruped robots moving through vineyard rows. It will allow for precise robot placement within the vineyard that is ideal for a robotic workspace. This allows the robot to perform selective, plant-by-plant task automation within the vineyard. A series of experiments were preformed with the Aliengo robot and our approach achieved a mean of 3.36cm and standard deviation of 2.19cm of distance from the desired position, which is sufficient for an automated task.

\section{State of the Art}
As of today, different robots and vehicles have been developed that can move autonomously throughout vineyards. These robots either move continuously throughout the row and/or are not quadrupeds. The EU Project BACCHUS robot is a wheeled vehicle that is under development to harvest grapes and take care of vineyards. The BACCHUS robot uses semantic segmentation of vineyard trunks for its localization \cite{loop_closure}. Our proposed navigation architecture takes the same segmentation approach but it is used to identify positions for the robot to walk to instead.  The EU Project CANOPIES is aimed at developing a human-robot collaborative paradigm for harvesting and pruning in vineyards\footnote[1]{www.canopies-project.eu}. It is a wheeled robot that works over the vineyard row. A similar autonomous over the vineyard row robot is the ViTiBOT Bakus which is used to improve vineyard help by removing herbicides and using precision spraying. This solution does not include stopping at each grapevine. YANMAR's autonomous over-the-row robot, YV01, does a similar task that autonomously sprays vineyard rows, without stopping at a specific grapevine\footnote[2]{https://www.yanmar.com/eu/campaign/2021/10/vineyard/}. A proposed wheeled robot for precision agriculture is the Agri.q02 which is meant to work in unstructured environments in collaboration with a UAV \cite{agriq}.  A navigation stack was created for the wheeled Ackerman Vehicles in percision farming, path planning from pose to pose \cite{ackerman}. There was autonomous navigation outlined in the Echord++ GRAPE experiment which maps a vineyard that uses a wheeled robot and moves to locations on the map to perform tasks \cite{echord_grape}.  These autonomous robots are all wheeled and most do not have to stop at precise locations in the vineyard. The proposed navigation architecture of this paper is quadruped navigation based on previous techniques used for localization to find precise positions for automated tasks to take place such as winter pruning and harvesting grapes.

\section{Navigation Architecture}
The navigation architecture is a combination of higher-level control and object detection.  The higher level control will make decisions on its movement path through a vineyard row based on the grapevine trunks detected.  The object detection was done by training a Mask-RCNN from Detectron2 \cite{wu2019detectron2}.

\subsection{Higher Level Control}
\begin{figure}[h]
\centering
\includegraphics[width=0.48\textwidth]{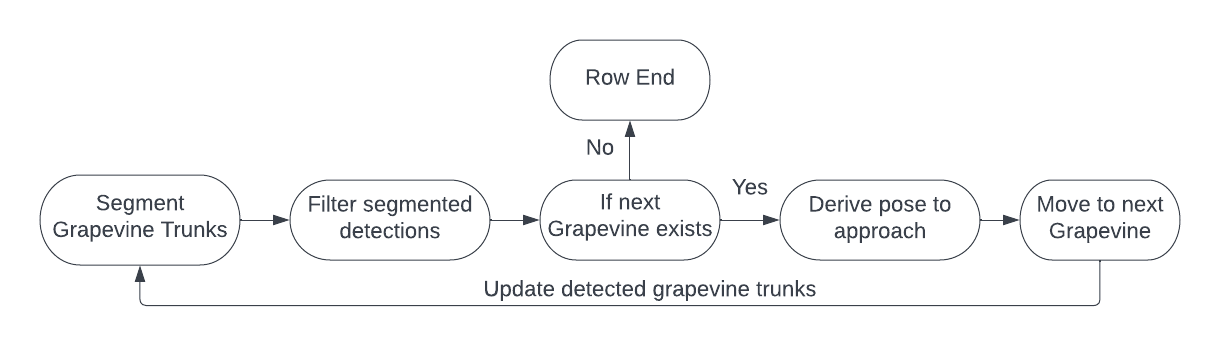}
\caption{Navigation Flow.}
\label{fig:Nav}
\end{figure}
The higher level control is a state machine for the robot to move throughout a vineyard row, as illustrated in Fig. \ref{fig:Nav}. It begins with an initial search to find the starting lines for both sides of the row. The user can set initially if they want the robot to move to the left or right of the row. The initial detections get sent through a filter which will find the rolling averages of each detection. From the filtered detection points, the control will find the lines on which the vineyard rows begin.

The robot has to approach parallel to the grapevines for it to be able to prune properly. To find the correct destination point, initially, the robot determines the orientation of the approach by calculating the vector of the vineyards in a row. This is derived from a list of points found in the initial search. It updates the vector for possible deviances of grapevines as the robot moves along the row.  The robot then approaches the grapevines in parallel at a desired distance that depends on the robot size and the workspace of the arm. 

After the robot has reached the determined location in the vineyard, it removes that grapevine from the list of vines to approach. Next, the control will choose the closest grapevine to the robot as its next target. It will continue this method until there are no more grapevines to identify in a row.

\subsection{Grapevine Identification}
Instance segmentation using a Mask-RCNN is used to detect the grapevine trunks in a vineyard. The training of the neural network was done in Detectron2 using 100 hand annotated images of potted grapevines. The corresponding depth of the detections is found by using the aligned depth image and from there the grapevine locations are found in relation to the quadruped.
\begin{figure}[h]
\centering
\captionsetup{justification=centering}
\includegraphics[width=0.40\textwidth]{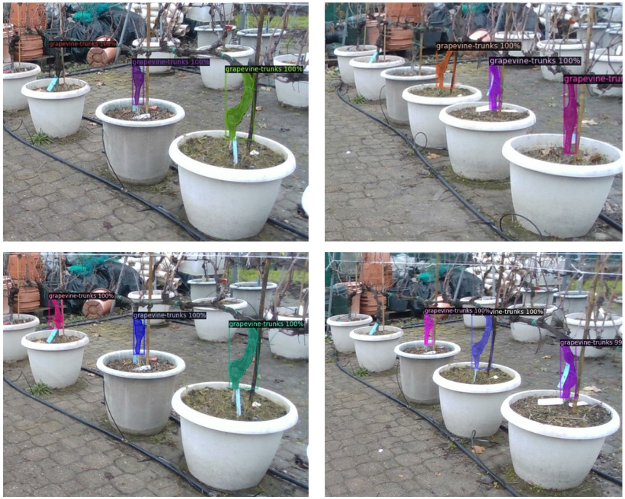}
\caption{Result of the image segmentation to detect grapevine trunks. (4 examples).}
\label{fig:trunk_det}
\end{figure}

\section{Experiments}
\subsection{Higher Level Control}

\subsubsection{Goals}
The goals of these experiments are to determine the precision of moving the robot's center of mass to desired positions. They are aimed to align the geometric center of the robot with the grapevine trunk, this way an arm mounted on the front of the robot is in the center of the grapevine's main cordon, and thus optimizes the workspace of the arm for single-plant operations, such as pruning.

\subsubsection{Setup}

\begin{figure}[h]
\centering
\includegraphics[width=0.48\textwidth]{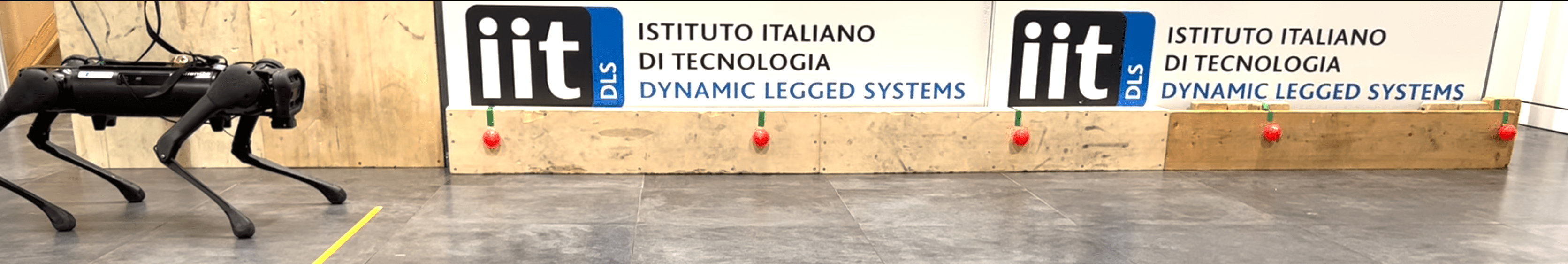}
\caption{Experiment setup.}
\label{fig:exp_setup}
\end{figure}

\begin{figure}[h]
\centering
\includegraphics[width=0.40\textwidth]{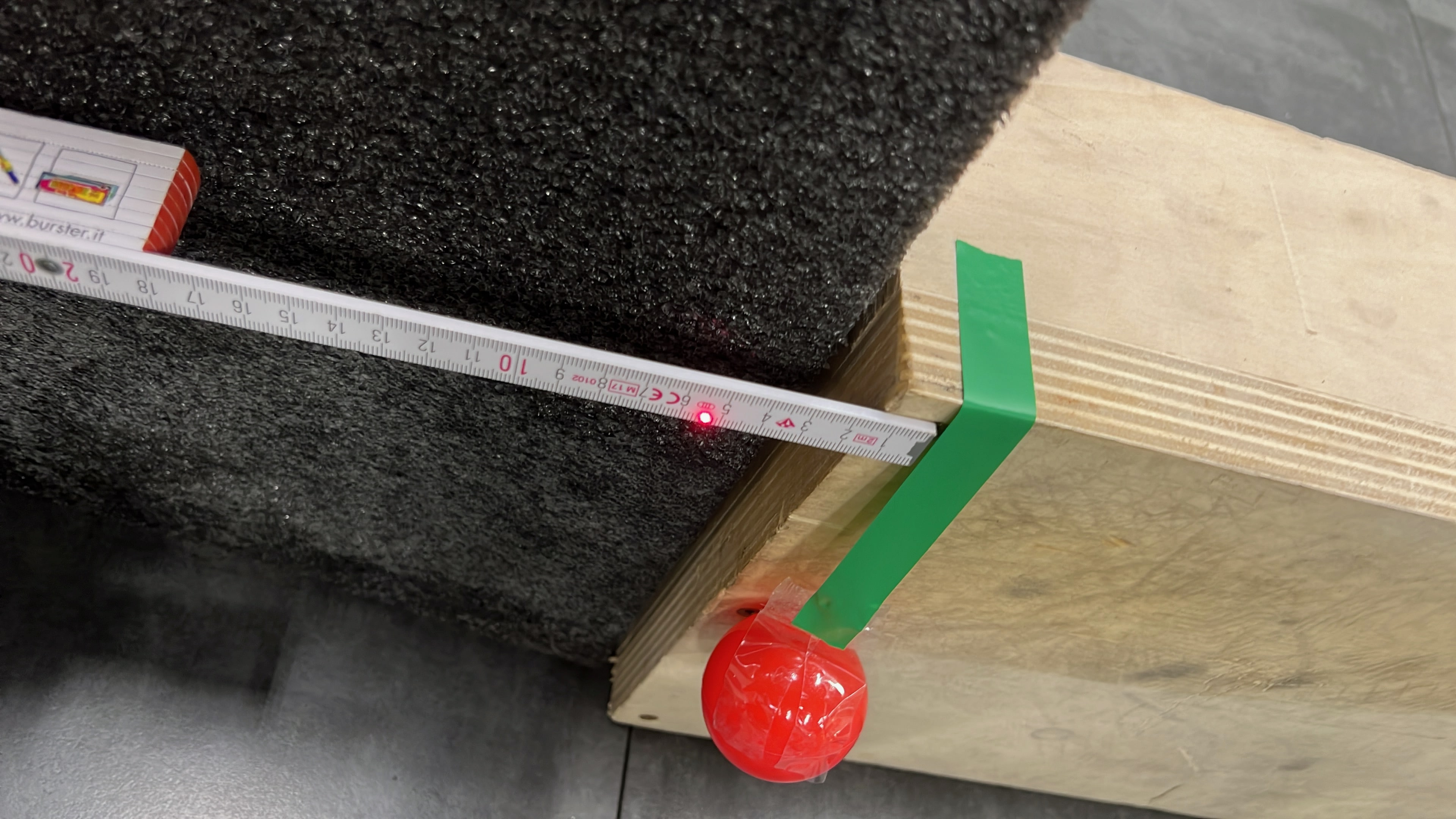} 
\caption{Measurement of Aliengo's arrival at a position.}
\label{fig:exp_measure}
\end{figure}

The higher level control was tested in a lab using Unitree’s Aliengo robot. Aliengo was used for simplification of experiments since it is 21kg and 61cm in length. Aliengo is equipped with Intel's Realsense D435 RGB-D camera. Red balls for segmenting were used to test in lab instead of the grapevine trunks. The red balls are spaced out at about 80cm from each other, the approximate distance that grapevines are from each other.
The setup of the experiment can be seen in Fig. \ref{fig:exp_setup}.  How the precision of the robot approaching a position was measured is shown in Fig. \ref{fig:exp_measure}.

\subsubsection{Tests}
 The robot does an initial search of the area using its RGB-D camera to segment the red balls. After it finds the row of red balls, it approaches the first position in the row. After the robot's arrival at the  initial position, it pauses for an automated task and update its detections. It repeats this process until the row is finished and then stops. 

Ten trials were conducted with five balls. To measure the error between the destination point and the red ball, a laser pointer was used to show the point that Aliengo's center of mass reached.

\subsubsection{Results} The error of reaching the destination point is a mean of 3.36cm and standard deviation of 2.19cm. The accompanying video shows complete trials.

\subsection{Grapevine Identification}
\subsubsection{Goals}
The goal of this is to test how well the Mask-RCNN was trained for working in vineyards.
\subsubsection{Setup}
The training of the neural network was done in Detectron2 using the framework set up in the paper \cite{DBLP:journals/corr/abs-2109-07247}.  
\subsubsection{Tests}
The results were tested on a previously recorded video of a potted vineyard at University Cattolica of Piacenza during winter.
\subsubsection{Results}
 Outputs from the model are shown in Fig. \ref{fig:trunk_det}. Currently the model needs to be trained on more data for robustness and for functionality in other vineyards as well.

\section{Conclusion}
This paper presented a method of computer-vision based navigation in vineyards for quadruped robots. This method will allow for precise placement to preform selective task automation.  

The control architecture works accurately with the experiments in the lab, and the trunk detections from the image segmentation can accurately identify grapevine trunks. The quadruped can reach a desired destination position with a mean error of 3.36cm error.  

The next steps for this architecture is combining the grapevine trunk semantic segmentation with the higher level control to test in the field. The dataset created for this project has to be expanded to train a more robust Mask-RCNN as well.

\section{Acknowledgments}
Thanks to the contributions of Miguel Fernandes for helping train the dataset and Lorenzo Amatucci for the configuration of the robot's controllers for experiments.

\printbibliography
\end{document}